# Ground Plane Detection with a Local Descriptor

Kangru Wang, Lei Qu, Lili Chen, Yuzhang Gu, Dongchen Zhu, Xiaolin Zhang

Shanghai Institute of Microsystem and Information Technology (SIMIT), Chinese Academy of Sciences (CAS)

*Abstrct* The detection of ground plane and free space remains challenging for non-flat plane, especially with the varying latitudinal and longitudinal slope or in the case of multi-ground plane. In this paper, we propose a framework of the ground plane detection with stereo vision. The main contribution of this paper is a newly proposed descriptor which is implemented in the disparity image to obtain a disparity texture image. The ground plane regions can be distinguished from their surroundings effectively in the disparity texture image. Because the descriptor is implemented in the local area of the image, it can address well the problem of non-flat plane. And we also present a complete framework to detect the ground plane regions base on the disparity texture image with convolutional neural network architecture.

*Index Terms* Ground Plane Detection, Free Space Estimation, Local Descriptor, Stereo Vision, Convolutional Network.

## I.INTRODUCTION

Ground plane and free space detection is a key component of intelligent vehicle applications and mobile robot system. The information of drivable space and understanding of the environment can improve traffic safety and efficiency. With the development of Driver Assistance Systems (ADAS) and Collision Avoidance Systems (CAS), frees pace detection has become an important research area of computer vision. Ground plane detection can be used for pitch angle compensation[[1] and improving the accuracy of obstacle detection [2], and free space estimation. Free space estimation is applied widely in many applications such as vehicle navigation [3] and pedestrian or vehicle detection [4, 5].

Labayrade et at[6] proposed the well-known 'V-disparity algorithm' which is a common approach for the ground plane modeling. The algorithm simplifies the extraction of the 3D ground and obstacle into a 2D linear process without using any prior knowledge of the scene appearance. The 'V-disparity' widely used to detect the ground plane [7-12]. [8]use a u-disparity map to get the detail information of ground. Traditional methods based on V-disparity map have some limitations in detecting non-flat ground especially in the off-road environment. Some more robust algorithms have been proposed to extract the non-flat ground plane .In[6] the authors assume that the ground plane can be modeled as piecewise linear and non-flat ground plane can be modeled by a few linear lines. [8, 13-15] address the problems with the ground plane of different longitudinal slope. But these methods usually fail in complex scene especially with wide variance of latitudinal slope and multi-ground plane. [18] use sliding window paradigm to address the detection of the ground plane with variable latitudinal slope and multi-ground plane. The plane is considered locally plane in very corresponding window, and sub-V-disparity map is created to represent the details of ground plane. But the number of window is hard to decide.

There are also some algorithms preserve the physical properties of the ground plane in Euclidian space[17-21]. [16]estimates ground plane using multivariate polynomial in the YZ plane domain .In[19,20],the input 3D map is reduced to a 2D map by accumulating all the points into a histogram of height versus distance which is similar to the v-disparity map creation. [21-23] applied a 2D quadratic surface fitting. [22, 23]introduces a method to estimate the planar patches for the Euclidian domain from the disparity map and then exploited the estimated patch parameters for eliminating outliers during road fitting. The traditional estimation of free space is usually based on the construction of occupancy grids [24, 25].The occupancy grid method models the occupancy evidence of the environment using a two-dimensional array or grid. Each cell of the occupancy grid maintains the probability of occupancy. However, these methods require the knowledge of the stereo sensor characteristics to compute depth map or 3D Euclidian point cloud as an initial step. With the development of deep learning, some algorithms to detect ground plane or free space have been proposed. [26] proposed a network to detect road that takes advantage of a large contextual window and uses a Network-in-Network (NiN) [27] proposed a multi-layer CNN architecture with a new loss function to detect free space.

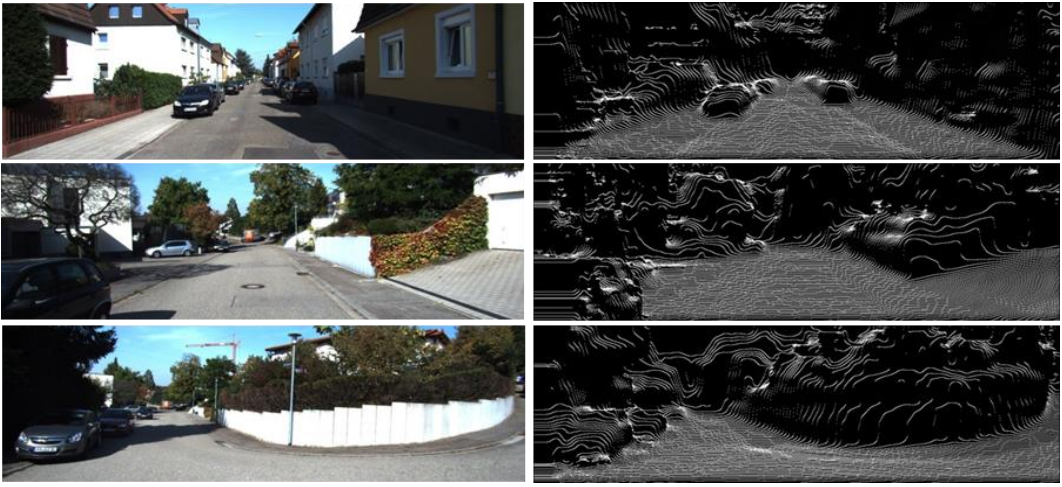

Figure 1: the left images are stereo left images and the right images are the disparity texture images.

In this paper, we propose a framework of the ground plane and free space estimation with stereo vision. The main contributions of this paper are presented as follows. First, we proposed a descriptor to obtain a texture map where the ground plane regions can be distinguished from their surroundings effectively. Second, the texture map is segmented into superpixel regions and use a convolutional neural network architecture to classify every superpixel region. We use the contextual information around the consider superpixel region to improve the accuracy of the model.

## II PROPOSED APPROACH

Our proposed method mainly consists of three steps, i.e. compute a disparity texture image, segment the disparity texture image, and detect the ground plane region using convolutional neural network architecture.

### A. Ground detection

*(1). Compute the disparity texture image*

In this paper, we propose a descriptor to extract ground and non-ground plane feature from disparity image. A disparity texture image can be obtained after computer every pixel in disparity image with the descriptor. This feature of the disparity texture image can distinguish ground plane regions from their surroundings effectively. This descriptor is implemented on the disparity image directly without using any other information. In this paper, the dense disparity estimation is performed using the algorithm of [28] with the reason of high quality.

Six typical planes in the world coordinate system are illustrated in Fig. 2(a) and their disparity maps are shown in Fig.2(b). The plane ① represents the horizontal ground plane, while the plane ② and the plane ③ represent the ground plane with latitudinal and longitudinal slope. The disparity value on the above typical ground plane should decrease gradually from the bottom to the top along vertical coordinate, while the disparity value keeps constant on the plane ①, ③ and decreases gradually on the plane ② along the horizontal coordinate. The plane ④, ⑤ and ⑥ represent the typical planes of obstacle, which have different disparity characteristics from the ground plane.

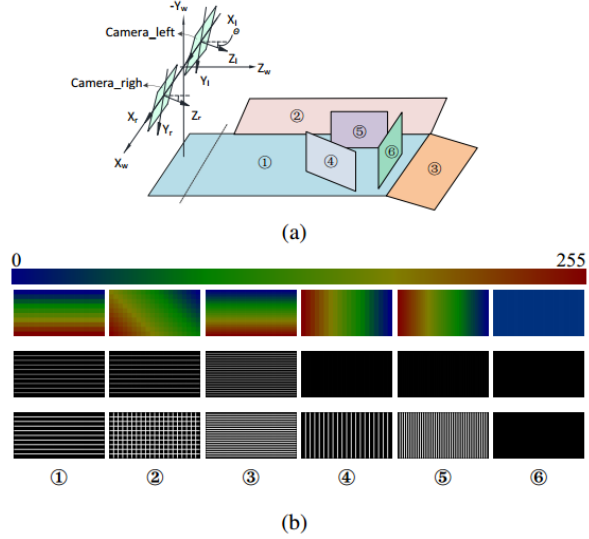

Fig. 2 Examples of the Disparity Texture Map. (a) Typical planes in world coordinate system. (b) The six images in the first row are the disparity maps of the planes in figure (a). The six images in the second row are the corresponding Disparity Texture Maps. The six images in the third row are the binary maps of the Disparity Texture Maps for visualization. The labels are consistent with the planes in figure (a). The color in the disparity maps encodes the disparity value. There are notable differences between the ground plane (plane ①, ②, ③) and obstacle (plane ④, ⑤, ⑥) in the Disparity Texture Maps.

Based on the disparity character of ground and non-ground plane,

*(2). Segment the disparity texture image*

The second step in our proposed framework is to segment the disparity texture map with the SLIC superpixel algorithm. Simple linear iterative clustering (SLIC) [29] is a good algorithm to generate superpixels by using spatial and color information with computational and memory efficiency. Ideally, the small region segmented by superpixel belongs to the same object. In addition, it shows excellent boundary adherence which can help improve the precision of segment between ground and non-ground region.

Thus, we utilize the SLIC superpixel algorithm to segment the stereo left image into superpixel regions. Then, rule of segment is used to divide the disparity texture image into corresponding superpixel regions. The feature extracted for each superpixel is used to determine the region class (ground or non-ground). We consider all the pixels in the superpixel region have the same class.

*(3). detect ground plane region*

In this step, we propose a convolutional neural network architecture to classify every superpixel region into ground or non-ground. It consists of extracting patches around superpixel regions of the disparity texture image and predicting the label of the superpixel using a trained CNN.

To reduce the impact of disparity estimation error and improve the accuracy of prediction, a possible solution is to make use of the contextual information around the considered superpixel region. Thus, in this framework we input the image patches centered at the centroid of each superpixel to our network and the output is a class (ground or non-ground) of the considered superpixel region.

Because the feature in disparity texture image is very obvious, we use a simple net architecture inspired by the LeNet-5 architecture which is also proposed to implement on gray image initially. The CN architecture is listed in Table I.

TABLE I

Convolutional Neural Network architecture

|    | Type of layer        | parameter |
|----|----------------------|-----------|
| 1  | Convolutional layer  | 5x5x20    |
| 2  | Non-linear           | Relu      |
| 5  | maximum              | 2x2       |
| 6  | Convolutional layer  | 3x3x20    |
| 7  | Non-linear           | Relu      |
| 10 | maximum              | 2x2       |
| 11 | Fully connected layer| O=500     |
| 12 | Non-linear           | Relu      |
| 13 | Fully connected layer| O=2       |

*B Road Segmentation*

(1) Road Segmentation based on convolutional neural network

The prior work usually detected road based on color information, which have a limited in the situations of extreme Illumination and road texture. To deal with above situations, we exploit the disparity texture map in the road detection.

In this paper, we exploit multiple modalities of data in road detection. We propose a Multi-Information network which takes a multi representation of disparity texture map and an RGB image as input. The approach makes the extraction of a feature set based on color and disparity information. To combine information from different features, we use late fusion in the network. The network architecture of our Multi-Information network is illustrated in Fig. 4, which is inspired by [26]. The input to the network is the patches of three-channel RGB image and disparity texture map and its output is a class (road or non-road) that is attributed to the 4×4 region in the center of the patches. To classify a whole image, a patch should be extracted and classified for every 4×4 region of the original image with a 4×4 stride). Each path of the network itself starts with a conv. 3×3 - 32 (32 filters sized 3×3 each) layer, followed by a conv. 1×1 - 16 and a max-polling layer of 2×2. These three layers are repeated in sequence with the same parameters. Finally, there is a fully-connected layer with 1000 neurons and a final layer with 2 neurons (one for each class). All convolutional layers have a stride of 1×1, are followed by the ReLU activation function and do not employ padding. The first fully connected layer is followed by the ReLU function while the final layer implements the Softmax loss function.

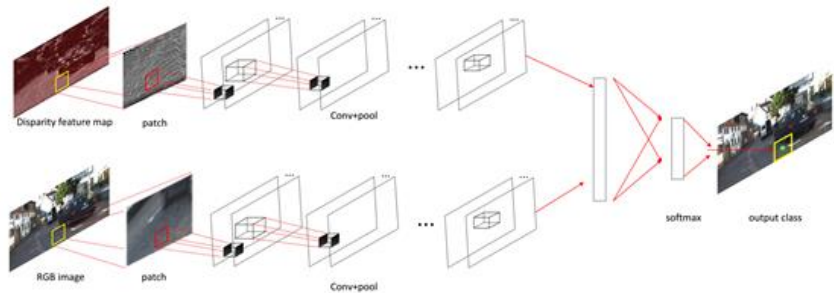

Fig. 4 The architecture of Multi-Information network.

*A. Dataset*

For the estimation experiments on ground detection we use the raw data available in the KITTI dataset [31]. The dataset consists of 122 images as training images and 303 as testing images, and ground truth is generated by manually. We choose the training images from 7 diverse sequences (09_26_d_09, 09_26_d_18, 09_26_d_35, 09_26_d_48, 09_28_d_34, 09_28_d_45, 09_28_d_68). and the testing images are chosen from other 10 diverse sequences (09_26_d_17, 09_26_d_39, 09_26_d_61, 09_26_d_64, 09_26_d_86, 09_26_d_93, 09_28_d_66, 09_30_d_33, 10_03_d_27, 10_03_d_47).This diverse sequences are obtained from different categories including city, residential, road, person and campus where terrains rang from plane to non-plane ground.

The experiment on road detection is evaluated on the datasets from the KITTI road benchmark [31]. The data are categorized in three sets having each one a subset of training and test images, representing a typical road scene in inner-city. The images in UU set are taken from urban unmarked area which have 98 training images and 100 testing images. The images in UM set are taken from urban marked two-way road which have 95 training images and 96 testing images. The images in UMM set are taken from urban marked multi-lane road, which have 96 training images and 94 testing images.

B. Ground detection

*(1)Training Scheme*

We created training and validation samples from the training images. Each sample consists of the image patch and its referent class. To create the samples, we extracted the image patch centered at the centroid of each superpixel region and the class (ground or non-ground) is attributed to the considered superpixel region. To make the balance of the samples, we sampled the non-ground class samples. We conduct the training using stochastic gradient descent.

*(2). Evaluation Result*

We compare our approach with two baselines: V-disparity [6] and Sub-V-disparity [18]. Fig.6 shows a sample of the obtained results. These demonstrate that our algorithm is able to provide superior average performance on non-flat ground. We also tested the effect of the size of block in our proposed descriptor on the model. The size of each block is 1x1, 3x3. A visual sample of the results can be seen in Fig. 7.

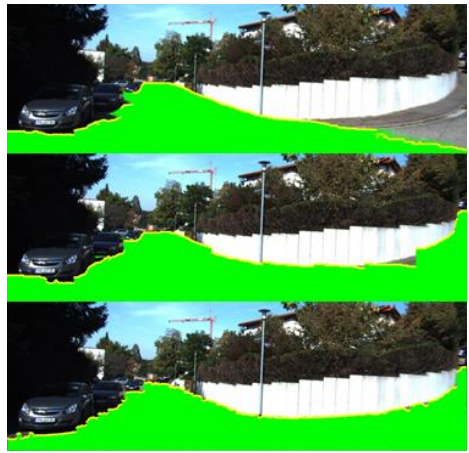

Figure 6 From the first to third column are the example results of v-disparity, sub-v-disparity and our method.

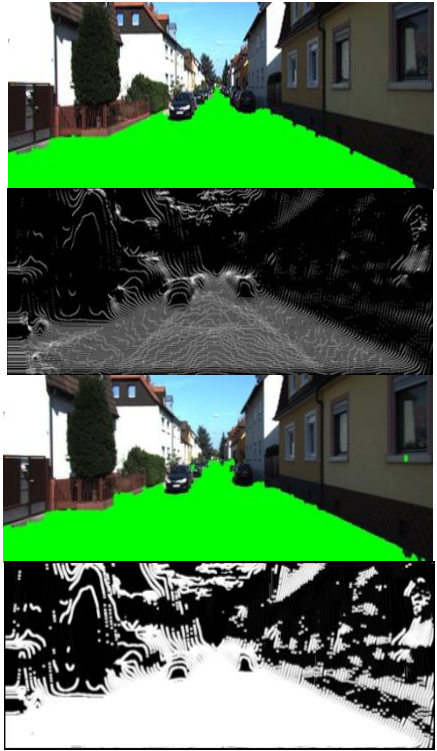

Figure 7 The images in the first and third columns are the ground plane detection results with the descriptors of 1x1 block size and 3x3 block size. The images in the second and fourth columns are the disparity texture images obtained by the descriptors of 1x1 block size and 3x3 block size.

C. Road Segmentation

*(1) Training Scheme*

Using the training images, we created training and validation datasets for each of the patch sizes tested. Each sample consists of the two patches of RGB image and disparity texture map. To create the samples, we scan the image skipping 4 pixels in each axis (stride of 4) and extracted the patch centered around each 4×4 region. We included only samples whose 4×4 regions are of a single class, ignoring ambiguous samples. All images are padded (using reflection) so the 4×4 regions cover the full original image.

*(2) Inference*

At inference time, our Multi-Information network is converted into a Fully Convolutional Network (FCNs) by converting the fully connected layers into convolutional layers. Concretely, we convert the two fully connected layer in our model to convolutional layers.

*(3) Evaluation Result*

We evaluate our approach with the approach proposed by [26], Fig 8. shows the qualitative results. In the situation of extreme Illumination and rare road texture, our approach provides more precise segmentation of the road than the baselines.

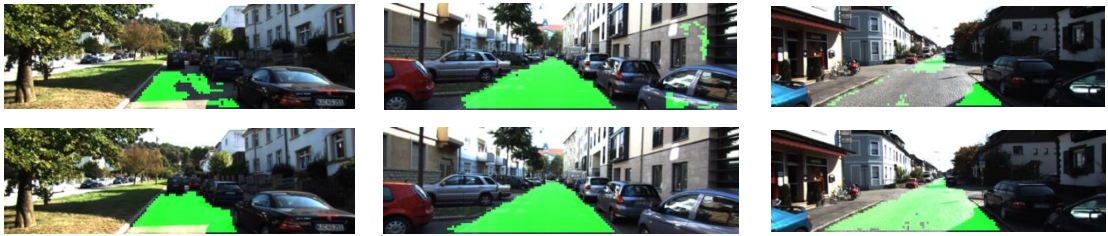

Figure 8: The images in first row are the result of approach [26]. The images in second row are the result of our approach.

## IV. CONCLUSION

We present a method of ground plane and free space estimation with stereo vision. We proposed a descriptor to obtain a disparity texture map where the ground plane regions can be distinguished from their surroundings effectively. A complete framework is proposed to detect the ground plane region base on the disparity texture image with a convolutional neural network architecture. The framework is shown to provide robust results over a variety of terrains from KITTI's benchmark. Our framework also benefits traditional methods with better results.